\documentclass{article}

\usepackage{microtype}
\usepackage{graphicx}
\usepackage{subfigure}
\usepackage{booktabs} 

\usepackage{hyperref}

\usepackage[accepted]{icml2020}

\icmltitlerunning{Stabilizing Differentiable Architecture Search via Perturbation-based Regularization}

\usepackage{amsmath}
\usepackage{mathrsfs}
\usepackage{color}
\usepackage{multirow}
\usepackage{threeparttable}
\usepackage{graphicx}
\newcommand{\tabincell}[2]{\begin{tabular}{@{}#1@{}}#2\end{tabular}}
\usepackage{adjustbox}
\usepackage{xspace}
\usepackage{caption} 
\usepackage{paralist}

\def\name{SDARTS\xspace}
\def\EXP{SDARTS-RS\xspace}
\def\ADV{SDARTS-ADV\xspace}

\begin{document}

\twocolumn[
\icmltitle{Stabilizing Differentiable Architecture Search via \\ Perturbation-based Regularization}



\icmlsetsymbol{equal}{*}

\begin{icmlauthorlist}
\icmlauthor{Xiangning Chen}{to}
\icmlauthor{Cho-Jui Hsieh}{to}
\end{icmlauthorlist}

\icmlaffiliation{to}{Department of Computer Science, UCLA}

\icmlcorrespondingauthor{Xiangning Chen}{xiangning@cs.ucla.edu}

\icmlkeywords{Machine Learning, ICML}

\vskip 0.3in
]



\printAffiliationsAndNotice{}  

\begin{abstract}
Differentiable architecture search (DARTS) is a prevailing NAS solution to identify architectures. Based on the continuous relaxation of the architecture space, DARTS learns a differentiable architecture weight and largely reduces the search cost. However, its stability has been challenged for yielding deteriorating architectures as the search proceeds. We find that the precipitous validation loss landscape, which leads to a dramatic performance drop when distilling the final architecture, is an essential factor that causes instability. Based on this observation, we propose a perturbation-based regularization - SmoothDARTS (SDARTS), to smooth the loss landscape and improve the generalizability of DARTS-based methods. In particular, our new formulations stabilize DARTS-based methods by either random smoothing or adversarial attack. The search trajectory on NAS-Bench-1Shot1 demonstrates the effectiveness of our approach and due to the improved stability, we achieve performance gain across various search spaces on 4 datasets. Furthermore, we mathematically show that SDARTS implicitly regularizes the Hessian norm of the validation loss, which accounts for a smoother loss landscape and improved performance.

\end{abstract}

\section{Introduction}

\begin{figure*}[!htb]
\centering
\subfigure[DARTS]{\label{fig:landscape_darts}\includegraphics[width=0.33\linewidth]{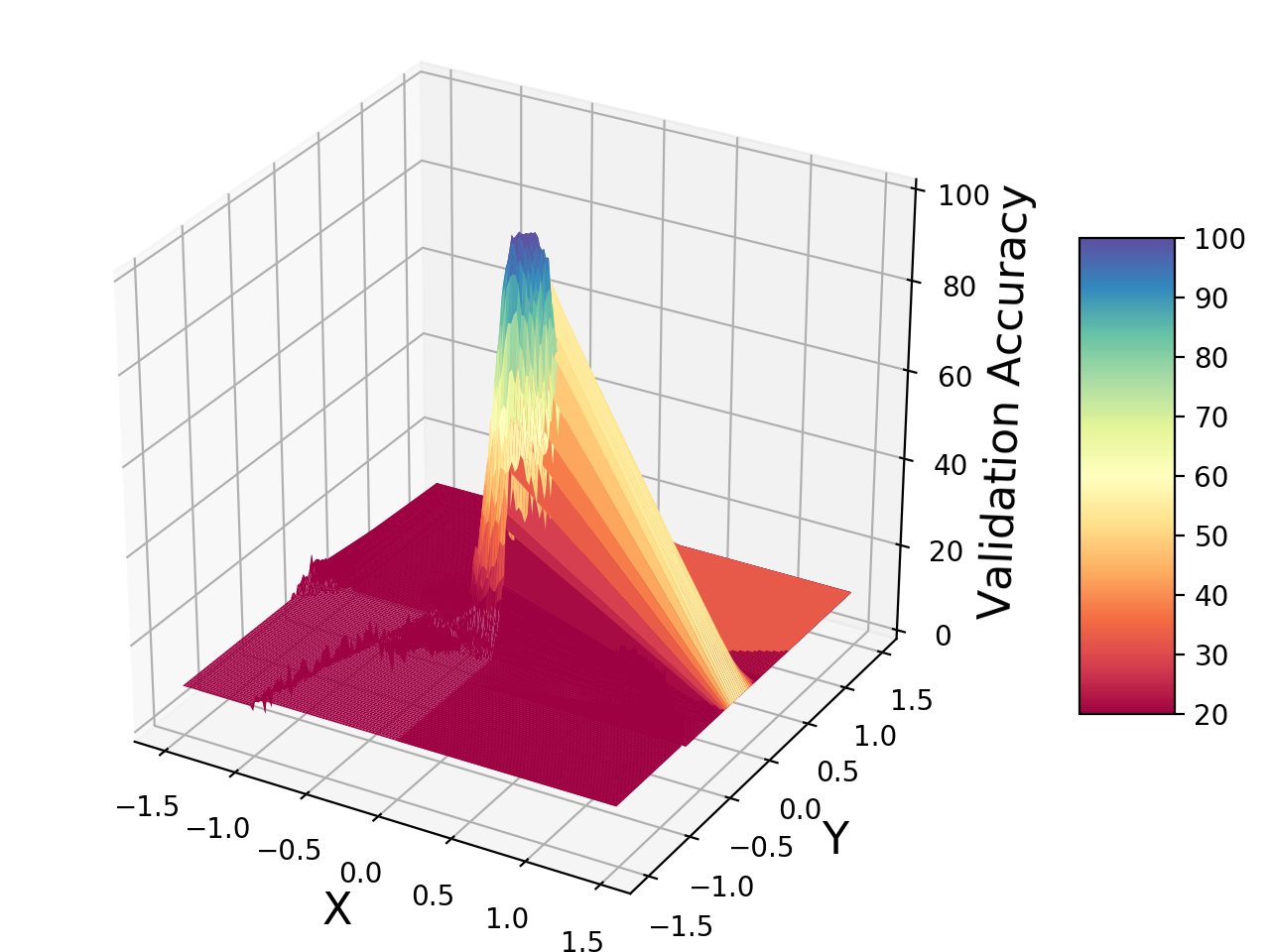}}
\subfigure[\EXP]{\label{fig:landscape_random}\includegraphics[width=0.33\linewidth]{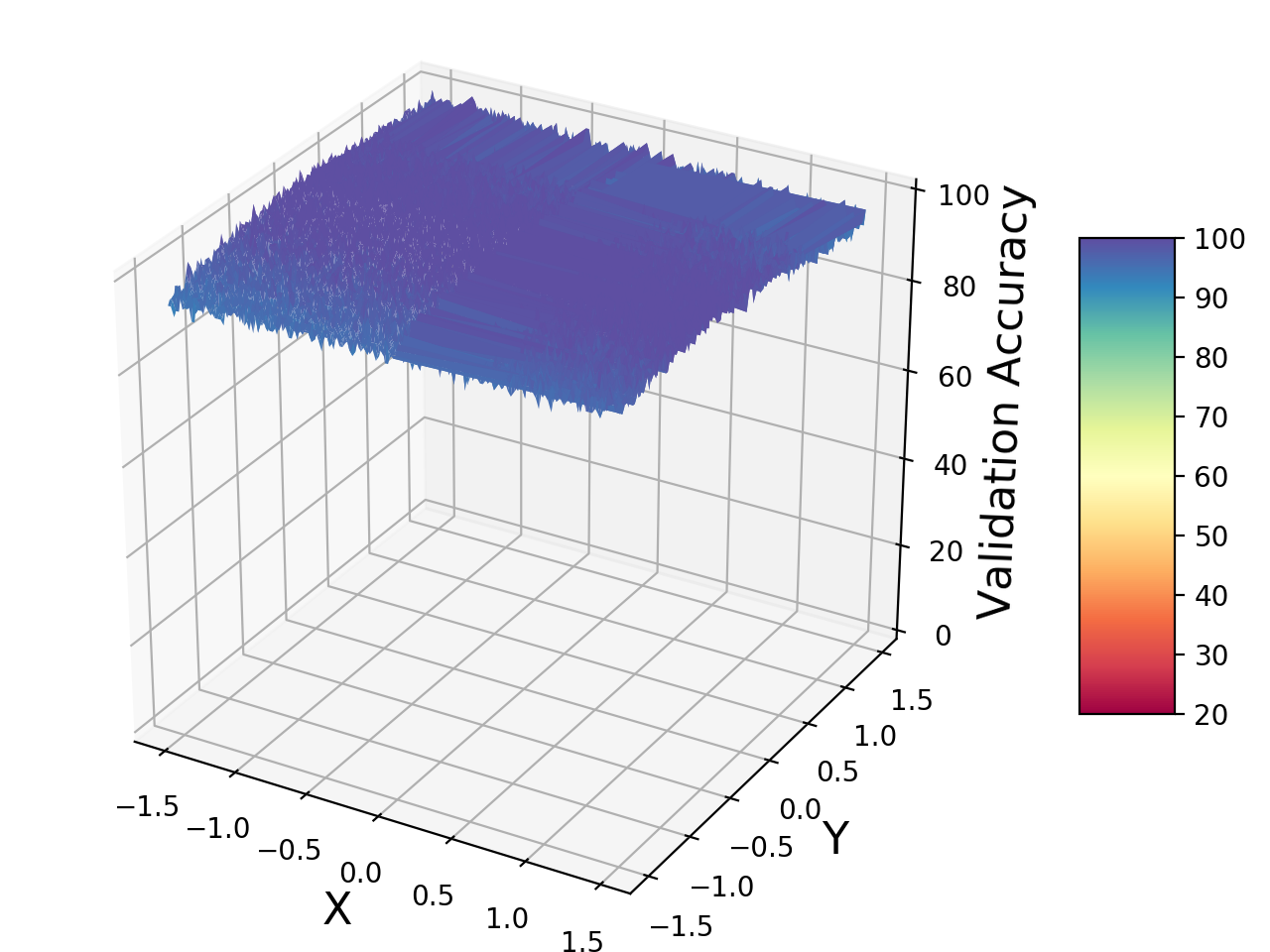}}
\subfigure[\ADV]{\label{fig:landscape_pgd}\includegraphics[width=0.33\linewidth]{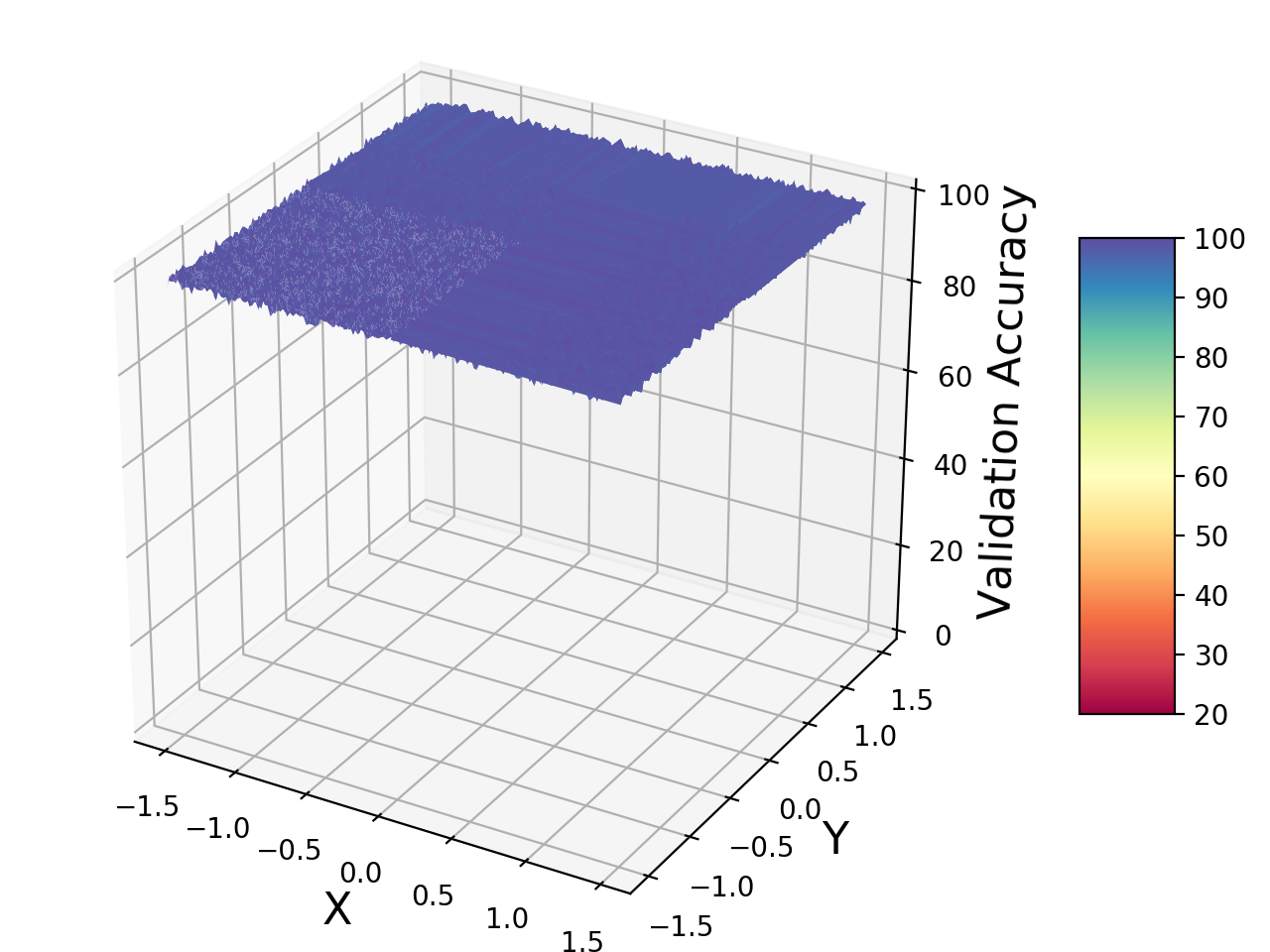}}
\vskip 0.1in
\caption{The landscape of validation accuracy regarding the architecture weight $A$ on CIFAR-10. The X-axis is the gradient direction $\nabla_{A}L_{valid}$, while the Y-axis is another random orthogonal direction (best viewed in color).}
\label{fig:landscape}
\end{figure*}
Neural architecture search (NAS) has emerged as a rational next step to automate the trial and error paradigm of architecture design. 
It is straightforward to search by reinforcement learning \cite{nas, nasnet, practicalnas} and evolutionary algorithm \cite{evolving, EvolvingDeep, LargeEvolution, HierarchicalEvo} due to the discrete nature of the architecture space.
However, these methods usually require massive computation resources. 
Recently, a variety of approaches are proposed to reduce the search cost including one-shot architecture search \cite{enas, oneshot, smash}, performance estimation \cite{LearningCP, accelerating} and network morphisms \cite{lamarckian, morphisms, pathlevel}. 
For example, one-shot architecture search methods construct a super-network covering all candidate architectures, where sub-networks with shared components also share the corresponding weights. Then the super-network is trained only once, which is much more efficient.
Based on this weight-sharing technique, DARTS \cite{darts} further builds a continuous mixture architecture and relaxes the categorical architecture search problem to learning a differentiable architecture weight $A$.

Despite being computationally efficient, the stability and generalizability of DARTS have been challenged recently. 
Many \cite{understanding, EvaluateSearch} have observed that although the validation accuracy of the mixture architecture keeps growing, the performance of the derived architecture collapses when evaluation. 
Such instability makes DARTS converge to distorted architectures. 
For instance, parameter-free operations such as \textit{skip connection} usually dominate the generated architecture \cite{understanding}, and DARTS has a preference towards wide and shallow structures \cite{WideShallow}. 
To alleviate this issue, \citet{understanding} propose to early stop the search process based on handcrafted criteria. 
However, the inherent instability starts from the very beginning and early stopping is a compromise without actually improving the search algorithm.

An important source of such instability is the final projection step to derive the actual discrete architecture from the continuous mixture architecture. 
There is often a huge performance drop in this projection step, so the validation accuracy of the mixture architecture, which is optimized by DARTS, may not be correlated with the final validation accuracy. 
As shown in Figure \ref{fig:landscape_darts}, DARTS often converges to a sharp region, so small perturbations will dramatically decrease the validation accuracy, let alone the projection step. 
Moreover, the sharp cone in the landscape illustrates that the network weight $w$ is almost only applicable to the current architecture weight $A$. 
Similarly, \citet{oneshot} also discovers that the shared weight $w$ of the one-shot network is sensitive and only works for a few sub-networks.
This empirically prevents DARTS from fully exploring the architecture space.


To address these problems, we propose two novel formulations.
Intuitively, 
the optimization of $A$ is based on $w$ that performs well on nearby configurations rather than exactly the current one. This leads to smoother landscapes as shown in Figure \ref{fig:landscape}(b, c).
Our contributions are as follows:

\begin{itemize}
\item 
We present \textbf{SmoothDARTS (SDARTS)} to overcome the instability and lack of generalizability of DARTS. 
Instead of assuming the shared weight $w$ as the minimizer with respect to the current architecture weight $A$, we formulate $w$ as the minimizer of the {\bf R}andomly {\bf S}moothed function, defined as the expected loss within the neighborhood of current $A$. 
The resulting approach, called \EXP, requires scarcely additional computational cost but is surprisingly effective. 
We also propose a stronger formulation that forces $w$ to minimize the worst-case loss around a neighborhood of $A$, which can be solved by {\bf ADV}ersarial training. 
The resulting algorithm, called \ADV,  leads to even better stability and improved performance. 
\item 
Mathematically, we show that the performance drop caused by discretization is highly related to the norm of Hessian regarding the architecture weight $A$, which is also mentioned empirically in \cite{understanding}. Furthermore, we show that both our regularization techniques are implicitly minimizing this term, which explains why our methods can significantly improve DARTS throughout various settings. 
\item 
The proposed methods consistently improve DARTS-based methods and can match or improve state-of-the-art results on various search spaces of CIFAR-10, ImageNet, and Penn Treebank.
Besides, extensive experiments show that our methods outperform other regularization approaches on 3 datasets across 4 search spaces.
Our code is available at \url{https://github.com/xiangning-chen/SmoothDARTS}.

\end{itemize}


\begin{figure}[!htb]
\centering
\subfigure[\EXP]{\label{fig:random_normal}\includegraphics[width=0.49\linewidth]{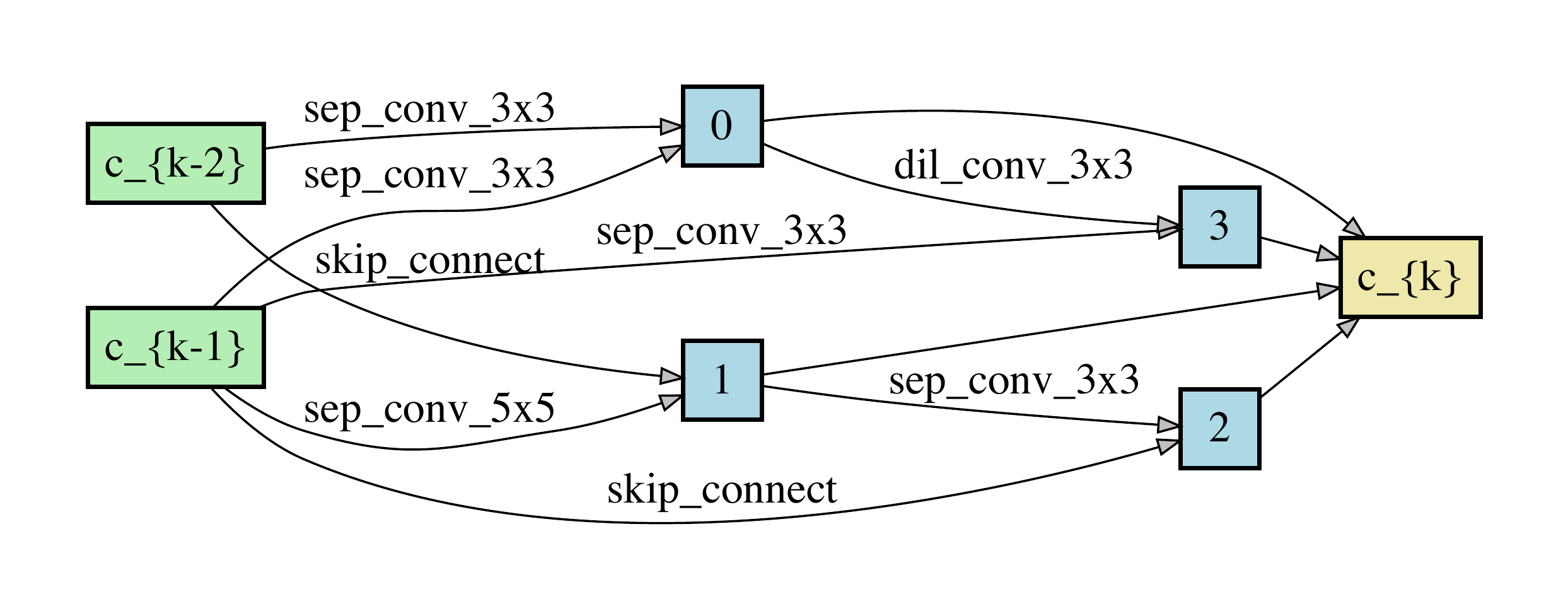}}
\subfigure[\ADV]{\label{fig:pgd_normal}\includegraphics[width=0.49\linewidth]{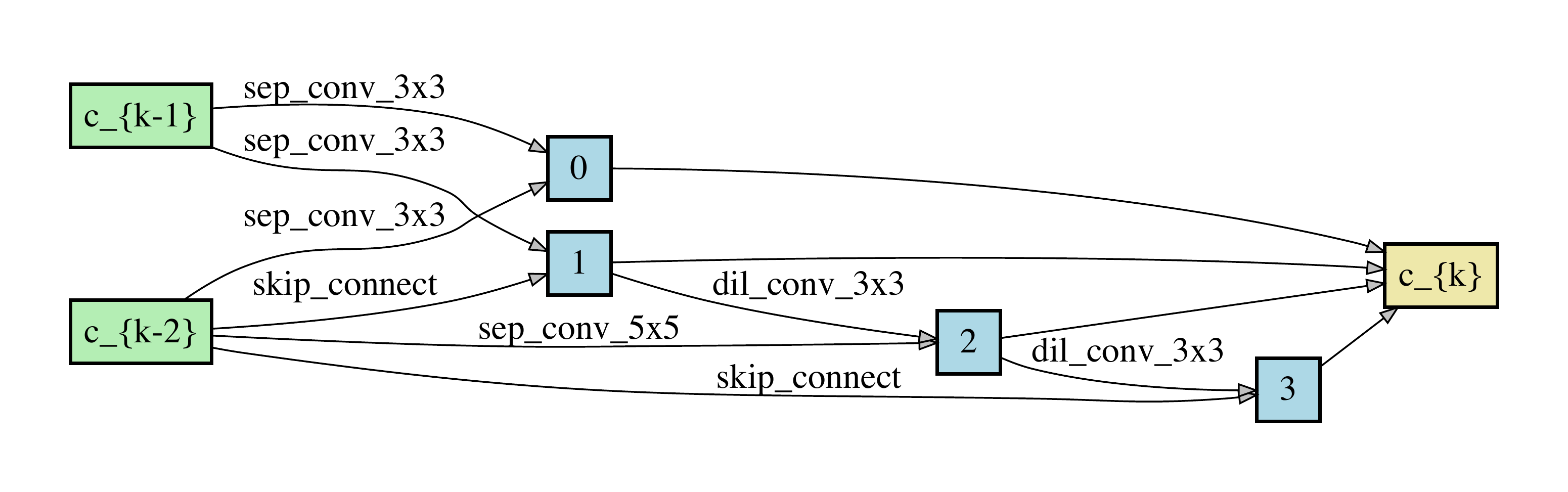}}
\caption{Normal cells discovered by \EXP and \ADV on CIFAR-10.}
\label{fig:normal}
\end{figure}

\section{Background and Related Work}

\subsection{Differentiable Architecture Search}

Similar to prior work \cite{nasnet}, DARTS only searches for the architecture of cells, which are stacked to compose the full network. 
Within a cell, there are $N$ nodes organized as a DAG (Figure \ref{fig:normal}), where every node $x^{(i)}$ is a latent representation and every edge $(i,j)$ is associated with a certain operation $o^{(i,j)}$.
It is inherently difficult to perform an efficient search since the choice of operation on every edge is discrete.
As a solution, DARTS constructs a mixed operation $\bar{o}^{(i,j)}$ on every edge:
\begin{equation}
    \bar{o}^{(i,j)}(x)=\sum_{o\in \mathcal{O}} \frac{\text{exp}(\alpha_{o}^{(i,j)})}
    {\sum_{o^{'}\in \mathcal{O}} \text{exp}(\alpha_{o^{'}}^{(i,j)})} o(x),  \nonumber
\end{equation}
where $\mathcal{O}$ is the candidate operation corpus and $\alpha_{o}^{(i,j)}$ denotes the corresponding architecture weight for operation $o$ on edge $(i,j)$.
Therefore, the original categorical choice per edge is parameterized by a vector $\alpha^{(i,j)}$ with dimension $|\mathcal{O}|$. 
And the architecture search is relaxed to learning a continuous architecture weight $A=[\alpha^{(i,j)}]$.
With such relaxation, DARTS formulates a bi-level optimization objective:
\begin{equation}
    \min_{A} L_{\text{val}}(w^*(A), A), \text{ s.t. } 
    w^* = \arg\min_{w} L_{\text{train}}(w, A).
    \label{eq:darts}
\end{equation}
Then, $A$ and $w$ are updated via gradient descent alternately, where $w^*$ is approximated by the current or one-step forward $w$.
DARTS sets up a wave in the NAS community and many approaches are springing up to make further improvements \cite{snas,gdas,nasp,pdarts,pcdarts,milenas,sif}.
For instance, PC-DARTS \cite{pcdarts} evaluates only a random proportion of channels during search, which can largely reduce the memory overhead. 
P-DARTS \cite{pdarts} attempts to narrow the gap between search and evaluation by progressively increasing the depth of the mixture architecture. 
Our regularization can be easily applied to these DARTS variants and bring consistent improvements.

\paragraph{Stabilize DARTS.} 
After search, DARTS simply prunes out operations on every edge except the one with the largest architecture weight when evaluation. 
Under such perturbation, its stability and generalizability have been widely challenged \cite{understanding, randomnas}.
\citet{understanding} empirically points out that the dominate eigenvalue $\lambda_{max}^{A}$ of the Hessian matrix $\nabla_{A}^{2}L_{valid}$ is highly correlated with the stability. 
They also present an early stopping criterion (DARTS-ES) to prevent $\lambda_{max}^{A}$ from exploding. 
Besides, partial channel connection \cite{pcdarts}, ScheduledDropPath \cite{nasnet} and L2 regularization on $w$ are also shown to improve the stability of DARTS.

\paragraph{NAS-Bench-1Shot1.}
NAS-Bench-1Shot1 is a benchmark architecture dataset \cite{nasbench1shot1} covering 3 search spaces based on CIFAR-10. 
It provides a mapping between the continuous space of differentiable NAS and discrete space in NAS-Bench-101 \cite{nasbench101} - the first architecture dataset proposed to lower the entry barrier of NAS. 
By querying in NAS-Bench-1Shot1, researchers can obtain necessary quantities for a specific architecture (e.g. test accuracy) in milliseconds. 
Using this benchmark, we track the anytime test error of various NAS algorithms, which allows us to compare their stability.


\subsection{Adversarial Robustness}
In this paper, we claim that DARTS should be robust against the perturbation on the architecture weight $A$. 
Similarly, the topic of adversarial robustness aims to overcome the vulnerability of neural networks against contrived input perturbation \cite{IntriguingProperty}.
Random smoothing \cite{lecuyer2019certified, cohen2019certified} is a popular method to improve model robustness.
Another effective approach is adversarial training \cite{fgsm, pgd}, which intuitively optimizes the worst-case training loss. In addition to gaining robustness, adversarial training has also been shown to improve the performance of image classification~\cite{xie2020adversarial} and GAN training~\cite{liu2019rob}. 
To the best of our knowledge, we are the first to apply this idea to stabilize the searching of NAS.


\section{Proposed method}
\subsection{Motivation}
During the DARTS search procedure, a continuous architecture weight $A$ is used, but it has to be projected to derive the discrete architecture eventually. 
There is often a huge performance drop in the projection stage, and thus a good mixture architecture does not imply a good final architecture. 
Therefore, although DARTS can consistently reduce the validation error of the mixture architecture, the validation error after projection is very unstable and could even blow up, as shown in Figure \ref{fig:NAS-Bench-1Shot1_explicit} and \ref{fig:NAS-Bench-1Shot1}. 

This phenomenon has been discussed in several recent papers \cite{understanding, darts+}, and \citet{understanding} empirically finds that the instability is related to the norm of Hessian $\nabla^2_A L_{\text{valid}}$. To verify this phenomenon, we plot the validation accuracy landscape of DARTS in Figure \ref{fig:landscape_darts}, which is extremely sharp -- small perturbation on $A$ can hugely reduce the validation accuracy from over 90\% to less than 10\%. 
This also undermines DARTS' exploration ability:
$A$ can only change slightly at each iteration because the current $w$ only works within a small local region.

\subsection{Proposed Formulation}


To address this issue, intuitively we want to force the landscape of $L_{\text{val}}(\bar{w}(A), A+\Delta)$ to be more smooth with respect to the perturbation $\Delta$. This leads to the following two versions of \name by redefining $\bar{w}(A)$: 
\begin{align}
\label{eq:proposed}
&\min_A L_{\text{val}}(\bar{w}(A), A), \text{ s.t. } \\
&\text{\EXP: } 
\bar{w}(A) = \arg\min_{w} E_{\delta \sim U_{[-\epsilon,\epsilon]}}
L_{\text{train}}(w, A+\delta) \nonumber\\
&\text{\ADV: }
\bar{w}(A) = \arg\min_{w} \max_{\|\delta\|\leq \epsilon }
L_{\text{train}}(w, A+\delta), \nonumber
\end{align}
where $U_{[-\epsilon, \epsilon]}$ represents the uniform distribution between $-\epsilon$ and $\epsilon$.
The main idea is that instead of using $w$ that only performs well on the current $A$, we replace it by the $\bar{w}$ defined in \eqref{eq:proposed} that performs well within a neighborhood of $A$. This forces our algorithms to focus on $(\bar{w}, A)$ pairs with smooth loss landscapes. 
For \EXP, we set $\bar{w}$ as the minimizer of the expected loss under small random perturbation bounded by $\epsilon$. 
This is related to   the idea of randomized smoothing, which randomly averages the neighborhood of a given function in order to obtain a smoother  and robust predictor \cite{cohen2019certified,lecuyer2019certified,liu2018towards,liu2020neural,li2019certified}.
On the other hand, we set $\bar{w}$ to minimize the worst-case training loss under small perturbation of $\epsilon$ for \ADV. This is based on the idea of adversarial training, which is a widely used technique in adversarial defense \cite{fgsm,madry2017towards}.


\begin{algorithm}[tb]
    \caption{Training of \name}
    \begin{algorithmic}
        \STATE Generate a mixed operation $\bar{o}^{(i,j)}$ for every edge $(i,j)$
        \WHILE{not converged}
            \STATE Update architecture $A$ by descending $\nabla_{A} L_{val}(w,A)$
            \STATE Compute $\delta$ based on equation \eqref{eq:darts-exp} or \eqref{eq:darts-adv}
            \STATE Update weight $w$ by descending $\nabla_w L_{train}(w, A+\delta)$
        \ENDWHILE
    \end{algorithmic}
    \label{alg:schema}
\end{algorithm}

\subsection{Search Algorithms}
The optimization algorithm for solving the proposed formulations is described in Algorithm \ref{alg:schema}. 
Similar to DARTS, our algorithm is based on alternating minimization between $A$ and $w$. 
For \EXP,  $\bar{w}$ is the minimizer of the expected loss altered by a randomly chosen $\delta$, which can be optimized by SGD directly.
We sample the following $\delta$ and add it to $A$ before running a single step of SGD on $w$
\footnote{We use uniform random for simplicity, while in practice this approach works also with other random perturbations, such as Gaussian. }:
\begin{align}
\label{eq:darts-exp}
\delta \sim U_{[-\epsilon, \epsilon]}.
\end{align}
This approach is very simple (adding only one line of the code) and efficient (doesn't introduce any overhead), and we find that it is quite effective to improve the stability. As shown in Figure \ref{fig:landscape_random}, the sharp cone disappears and the landscape becomes much smoother, which maintains high validation accuracy under perturbation on $A$.

For \ADV, we consider the worst-case loss under certain perturbation level, which is a stronger requirement than the expected loss in \EXP.
The resulting landscape is even smoother as illustrated in Figure~\ref{fig:landscape_pgd}.  
In this case, updating $\bar{w}$ needs to solve a min-max optimization problem beforehand. 
We employ the widely used multi-step projected gradient descent (PGD) on the negative training loss to iteratively compute $\delta$:
\begin{align}
\label{eq:darts-adv}
\delta^{n+1} = \mathcal{P}(\delta^n + lr*\nabla_{\delta^n}L_{\text{train}}(w, A+\delta^n))
\end{align}
where $\mathcal{P}$ denotes the projection onto the chosen norm ball (e.g. clipping in the case of the $\ell_{\infty}$ norm) and $lr$ denotes the learning rate.

In the next section, we will mathematically explain why \EXP and \ADV improve the stability and generalizability of DARTS. 

\section{Implicit Regularization on Hessian Matrix}\label{sec:RegHessian}
It has been empirically pointed out in \cite{understanding} that the dominant eigenvalue of $\nabla^2_{A}L_{\text{val}}(w, A)$ (spectral norm of Hessian) is highly correlated with the generalization quality of DARTS solutions. In standard DARTS training, the Hessian norm usually blows up, which leads to deteriorating (test) performance of the solutions. 
In Figure \ref{fig:NAS-Bench-1Shot1_eigen}, we plot this Hessian norm during the training procedure and find that the proposed methods, including both \EXP and \ADV, consistently reduce the Hessian norms during the training procedure.
In the following, we first explain why the spectral norm of Hessian is correlated with the solution quality, and then formally show that our algorithms can implicitly control the Hessian norm.

\paragraph{Why is Hessian norm correlated with solution quality?}
Assume $(w^*, A^*)$ is the optimal solution of \eqref{eq:darts} in the continuous space while $\bar{A}$ is the discrete solution by projecting $A^*$ to the simplex. Based on Taylor expansion and assume $\nabla_A L_{\text{val}}(w^*, A^*)= 0$ due to optimality condition, we have
\begin{equation}
L_{\text{val}}(w^*, \bar{A}) \!=\!
L_{\text{val}}(w^*, A^*) \!+\! \frac{1}{2} (\bar{A}-A^*)^T  \bar{H} (\bar{A}-A^*),
\label{eq:taylor}
\end{equation}
where $\bar{H} = \int_{A^*}^{\bar{A}} \nabla^2_A L_{\text{val}}(w^*, A) dA $ is the average Hessian. 
If we assume that Hessian is stable in a local region,
then the quantity of $C  = \|\nabla^2_{A} L_{\text{val}}(w^*, A^*)\|\|\bar{A}-A^*\|^2$ can approximately bound the performance drop when projecting $A^*$ to $\bar{A}$ with a fixed $w^*$. 
After fine tuning, $L_{\text{val}}(\bar{w}, \bar{A})$ where $\bar{w}$ is the optimal weight corresponding to $\bar{A}$ is expected to be even smaller than $L_{\text{val}}(w^*, \bar{A})$, if the training and validation losses are highly correlated. 
Therefore, the performance of $L_\text{val}(\bar{w}, \bar{A})$, which is the quantity we care, will also be bounded by $C$. 
Note that the bound could be quite loose since it assumes the network weight remains unchanged when switching from $A^*$ to $\bar{A}$.
A more precise bound can be computed by viewing $g(A) = L_{\text{val}}(w^*(A), A)$ as a function only paramterized by $A$, and then calculate its derivative/Hessian by implicit function theory. 

\paragraph{Controlling spectral norm of Hessian is non-trivial. }

\begin{figure}[!htb]
\centering
\resizebox{.3\textwidth}{!}{
\includegraphics[width=0.6\linewidth]{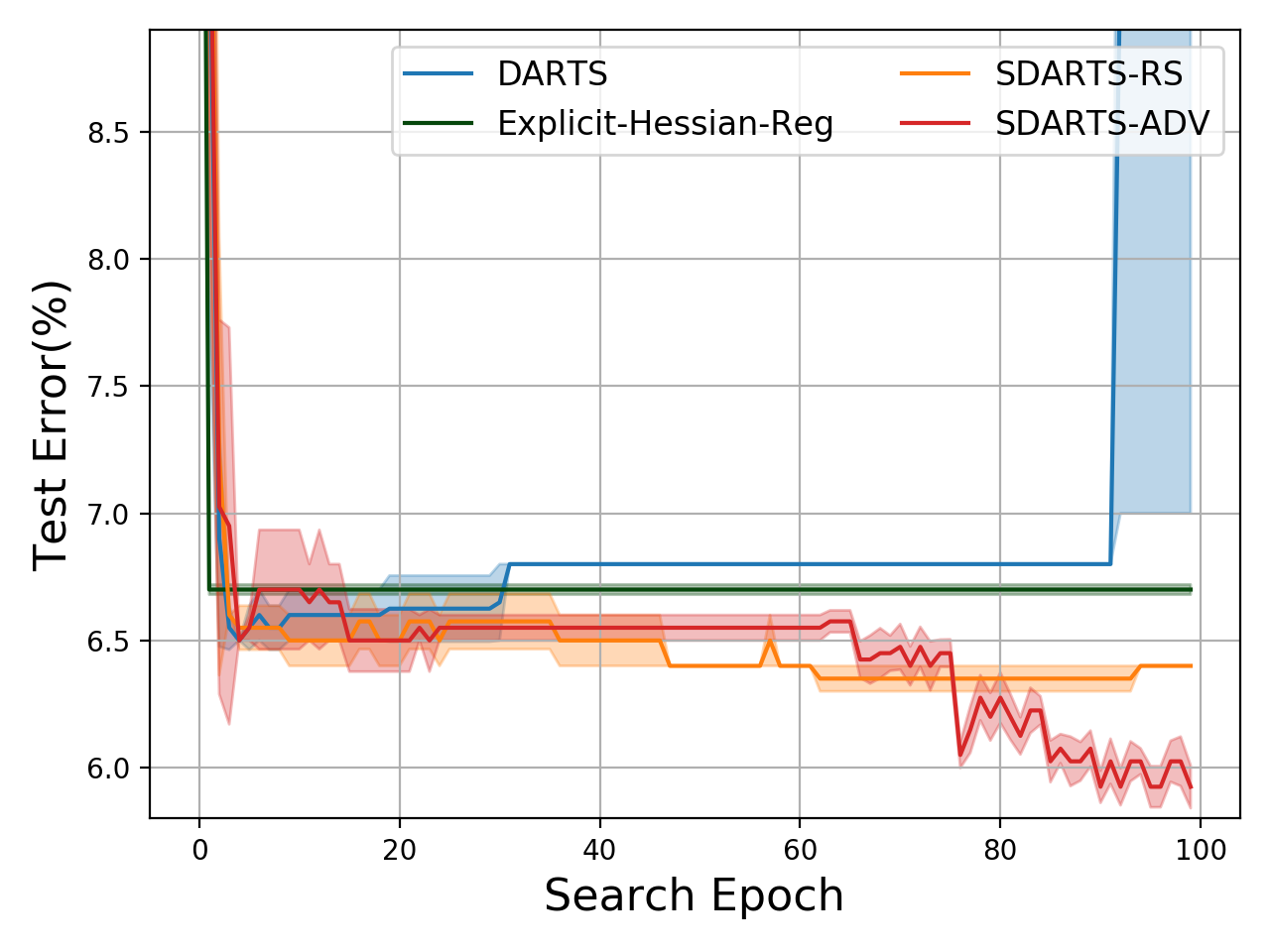}}
\caption{Anytime test error (mean $\pm$ std) of DARTS, explicit Hessian regularization, \EXP and \ADV on NAS-Bench-1Shot1 (best viewed in color).}
\label{fig:NAS-Bench-1Shot1_explicit}
\end{figure}

With the observation that the solution quality of DARTS is related to $\|\nabla_A^2 L_{\text{val}}(w^*, A^*)\|$, an immediate thought is to explicitly control this quantity during the optimization procedure. To implement this idea, we add an auxiliary term - the finite difference estimation of Hessian matrix
$(\nabla_{A}L_{\text{val}}(A+\epsilon)- \nabla_{A}L_{\text{val}}(A-\epsilon))/2\epsilon$ to the loss function when updating $A$. However, this requires much additional memory to build a computational graph of the gradient, and Figure \ref{fig:NAS-Bench-1Shot1_explicit} suggests that it takes some effect compared with DARTS but is worse than both \EXP and \ADV. 
One potential reason is the high dimensionality -- there are too many directions of $\epsilon$ to choose from and we can only randomly sample a subset of them at each iteration.



\paragraph{Why can \EXP implicitly control Hessian?}\mbox{}\\
In \EXP, the objective function becomes
\begin{flalign}
    &E_{\delta \sim U_{[-\epsilon,\epsilon]}} L(w, A+\delta) \nonumber\\
    \approx
    &E_{\delta \sim U_{[-\epsilon,\epsilon]}} \bigg[ L(w, A) + \delta \nabla_A L(w, A) + 
    \frac{1}{2} \delta^T \nabla^2_A L(w, A) \delta \bigg]
    \label{eq:ii}\\
    =& L(w, A) + \frac{\epsilon^2}{6} \text{Tr}\bigg\{ \nabla^2_A L(w, A) \bigg\}, \nonumber
\end{flalign}
where the second term in \eqref{eq:ii} is canceled out since $E[\delta]=0$ and the off-diagonal elements of the third term becomes $0$ after taking the expectation on $\delta$.
The update of $w$ in \EXP can thus implicitly controls the trace norm of $\nabla^2_{A} L(w, A)$.  If the matrix is close to PSD, this is approximately regularizing the (positive) eigenvalues of $\nabla^2_{A}L_{\text{val}}(w, A)$.  Therefore, we observe that \EXP reduces the Hessian norm through its training procedure. 
 
\paragraph{Why can \ADV implicitly control Hessian?}
\ADV ensures that the validation loss is small under the worst-case perturbation of $A$.
If we assume the Hessian matrix is roughly constant within $\epsilon$-ball, then adversarial training implicitly minimizes
\begin{align}
    &\min_{A: \|A-A^*\|\leq \epsilon}
    L(w, A) \label{eq:aaa}\\
    \approx & L(w, A^*) + 
    \frac{1}{2}
    \max_{\|\Delta\|\leq \epsilon}
    \Delta^T H \Delta 
    \label{eq:bbb}
\end{align}
when the perturbation is in $\ell_2$ norm, the second term becomes the $\frac{1}{2} \epsilon^2 \|H\|$, 
and when the perturbation is in $\ell_{\infty}$ norm, the second term is bounded by $\epsilon^2 \|H\|$.
Thus \ADV also approximately minimizes the norm of Hessian.
In addition, notice that from \eqref{eq:aaa} to \eqref{eq:bbb} we assume the gradient is $0$, which is the property holds only for $A^*$. 
In the intermediate steps for a general $A$, the stability under perturbation will not only be related to Hessian but also gradient, 
and in \ADV we can still implicitly control the landscape to be smooth by minimizing the first-order term in the Taylor expansion of \eqref{eq:aaa}.

\section{Experiments}\label{sec:experiments}
In this section, we first track the anytime performance of our methods on NAS-Bench-1Shot1 in Section \ref{sec:nas-bench-1shot1}, which demonstrates their superior stability and generalizability. 
Then we perform experiments on the widely used CNN cell space with CIFAR-10 (Section \ref{sec:cifar10}) and ImageNet (Section \ref{sec:imagenet}). 
We also show our results on RNN cell space with PTB (Section \ref{sec:rnn_standard}). 
In Section \ref{sec:compare_reg}, we present a detailed comparison between our methods and other popular regularization techniques.
At last, we examine the generated architectures and illustrate that our methods mitigate the bias for certain operations and connection patterns in Section \ref{sec:examine_arch}.

\subsection{Experiments on NAS-Bench-1Shot1}
\label{sec:nas-bench-1shot1}
\begin{figure*}[!htb]
\centering
\resizebox{0.96\textwidth}{!}{
\subfigure[Space 1]{\label{fig:space1}\includegraphics[width=0.33\linewidth]{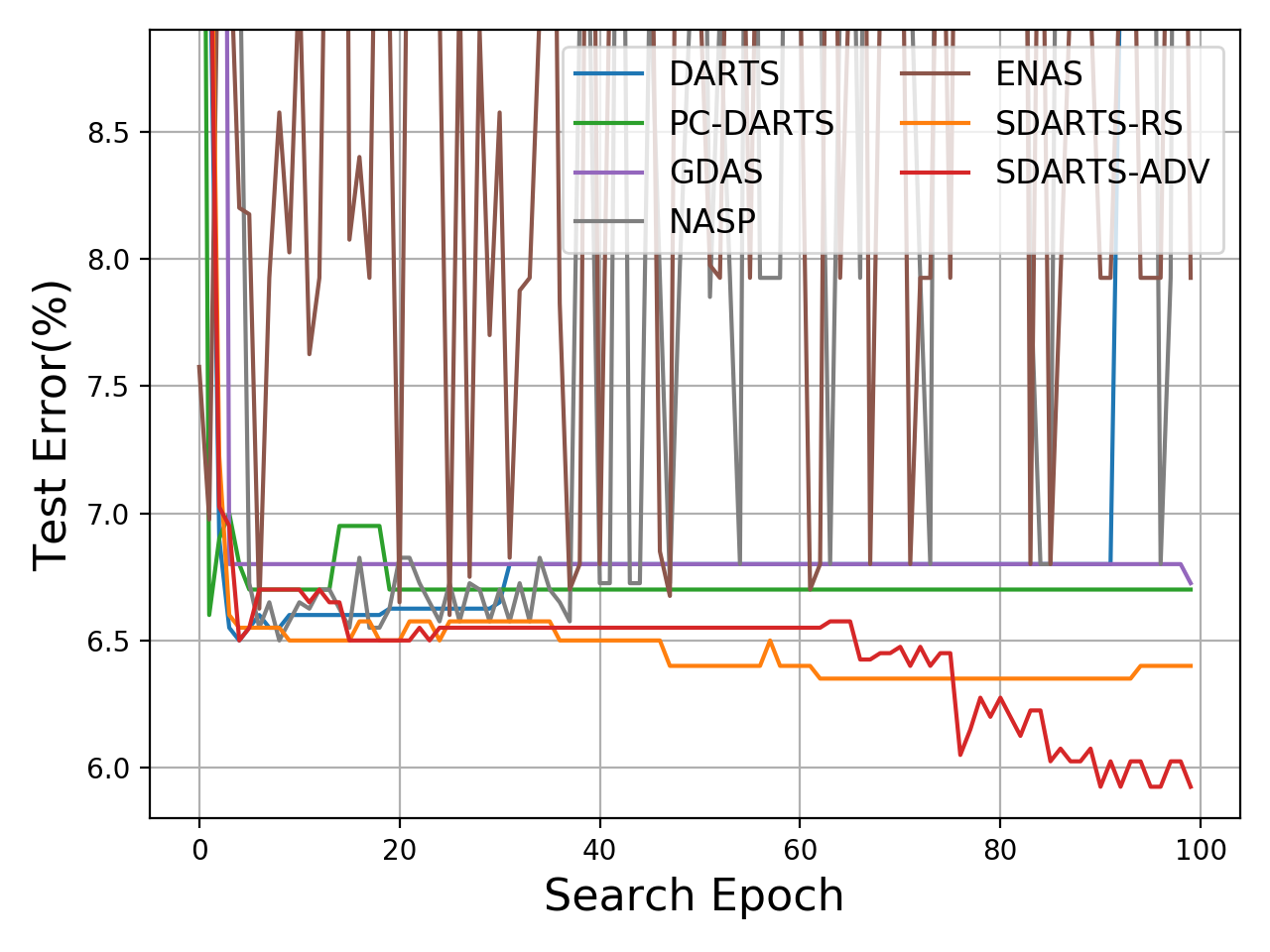}}
\subfigure[Space 2]{\label{fig:space2}\includegraphics[width=0.33\linewidth]{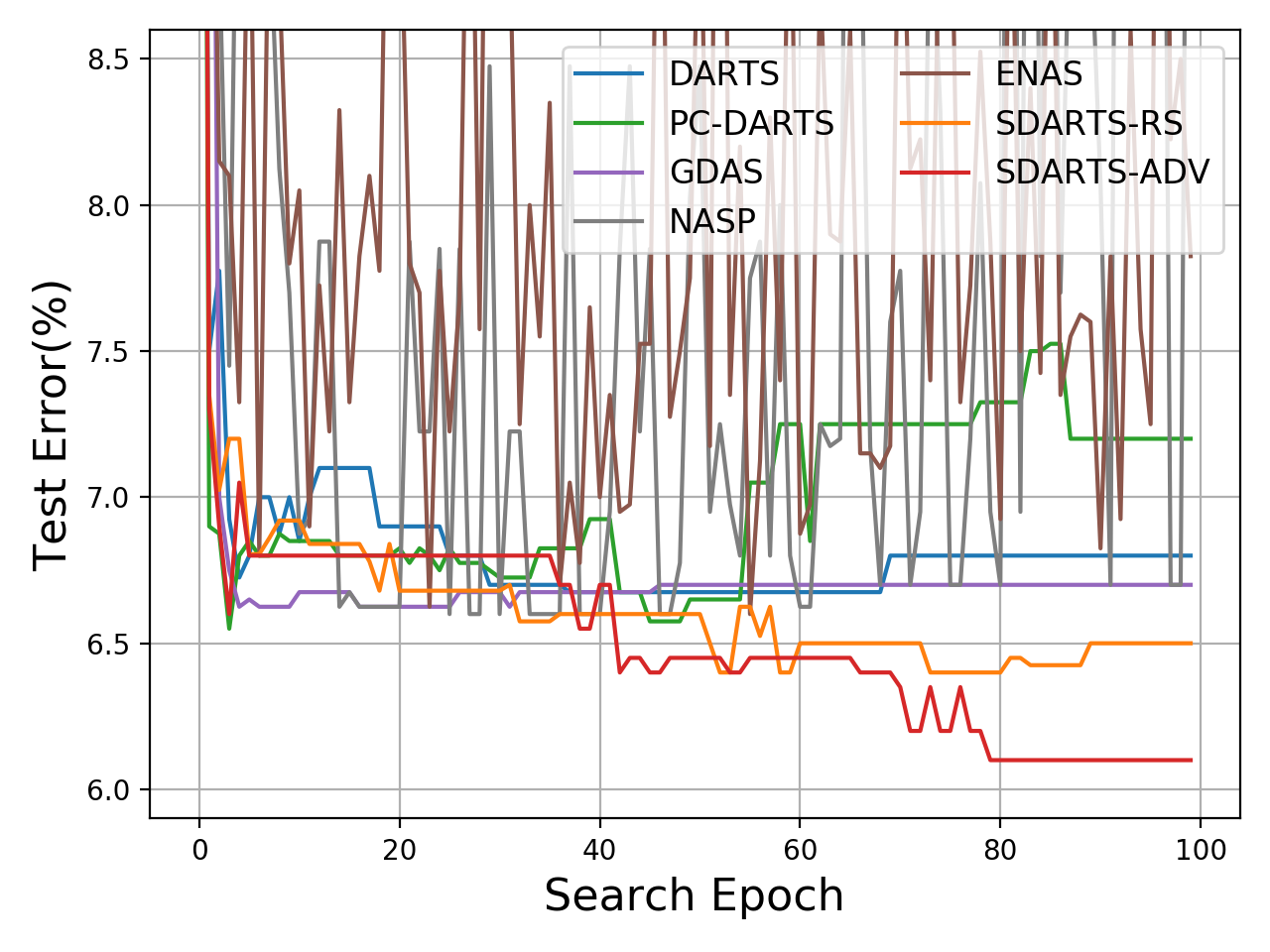}}
\subfigure[Space 3]{\label{fig:space3}\includegraphics[width=0.33\linewidth]{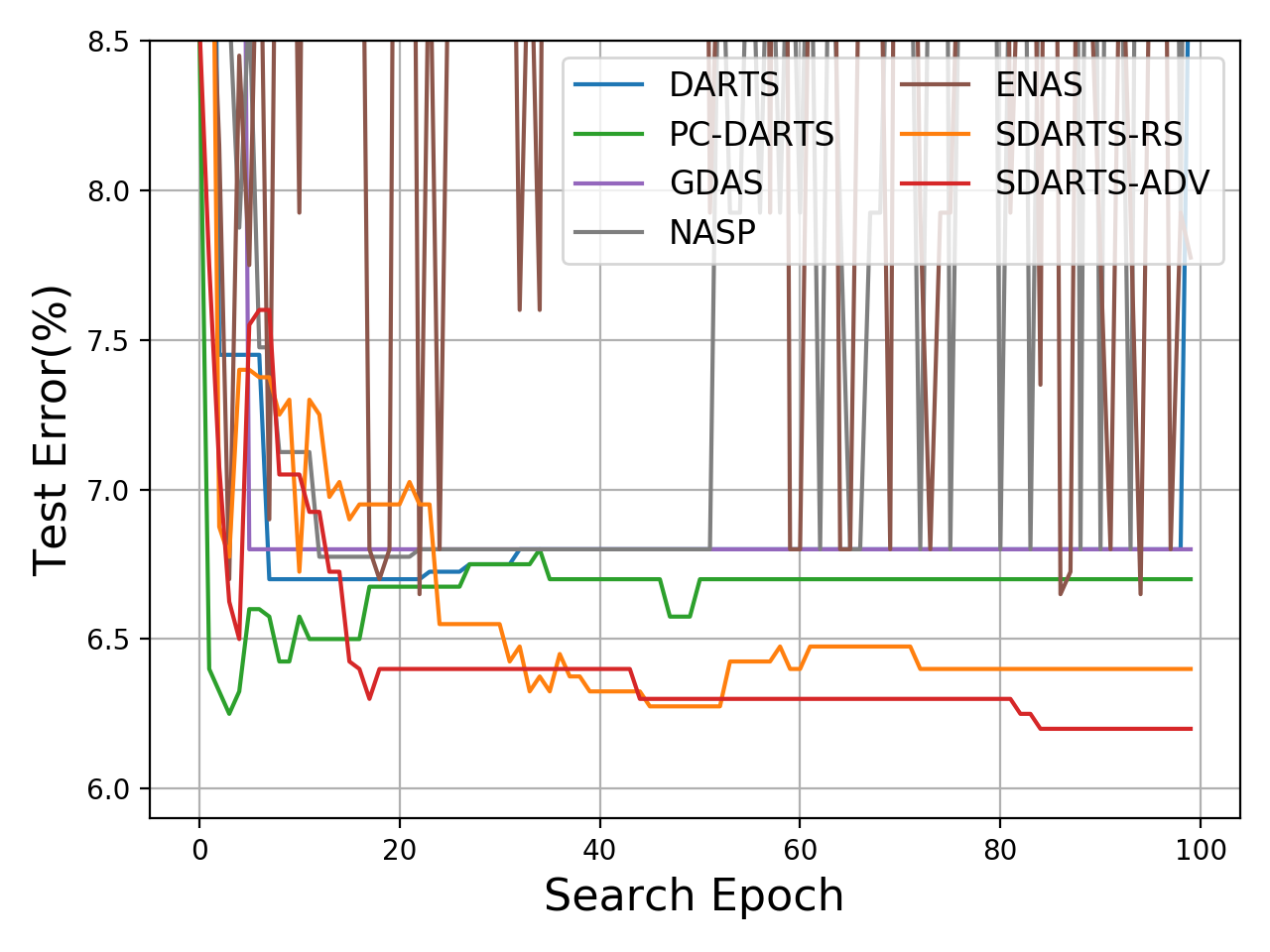}}
}
\caption{Anytime test error on NAS-Bench-1Shot1 (best viewed in color).}
\label{fig:NAS-Bench-1Shot1}
\end{figure*}

\begin{figure*}[!htb]
\centering
\resizebox{0.96\textwidth}{!}{
\subfigure[Space 1]{\label{fig:space1_eigen}\includegraphics[width=0.33\linewidth]{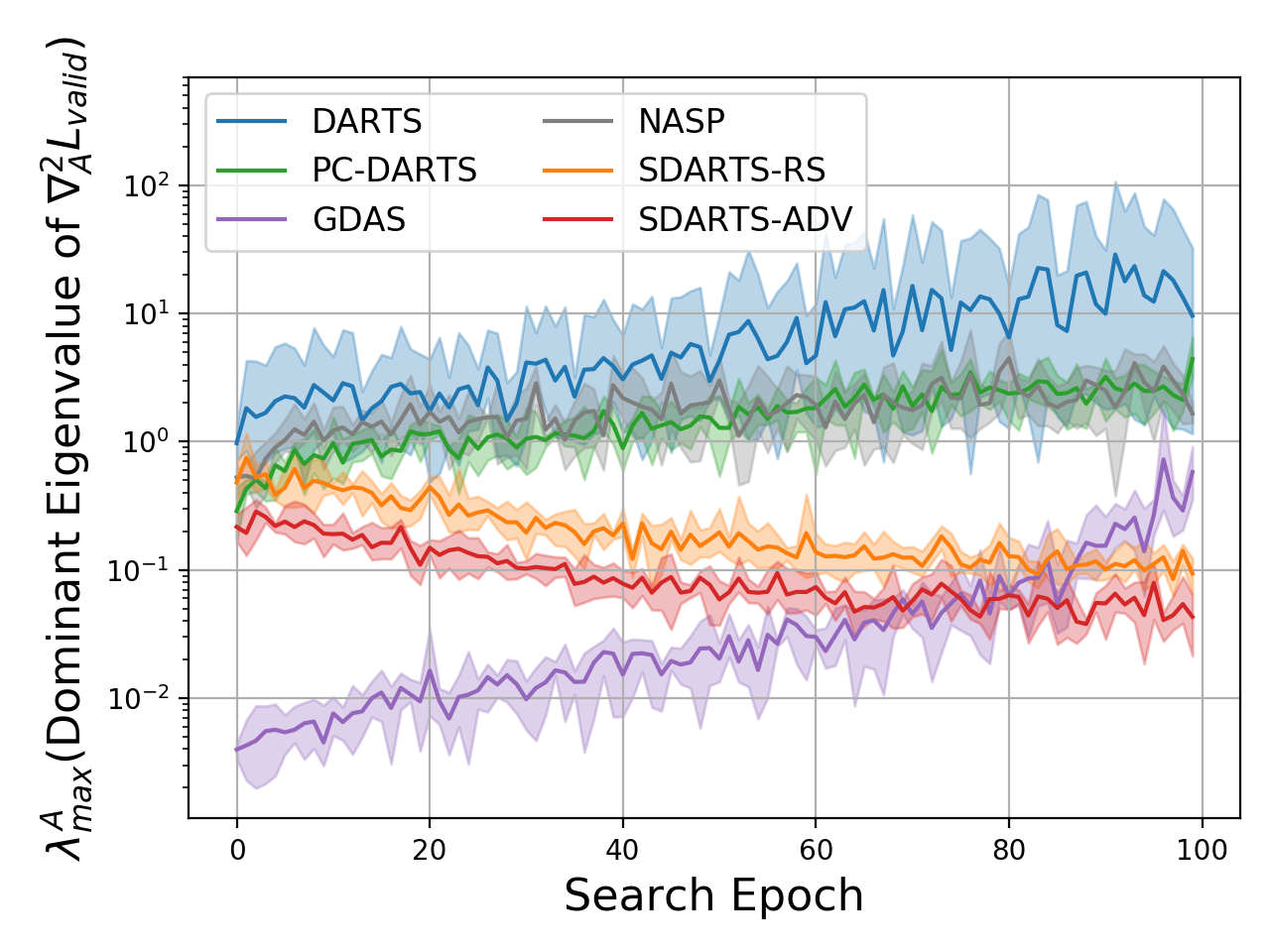}}
\subfigure[Space 2]{\label{fig:space2_eigen}\includegraphics[width=0.33\linewidth]{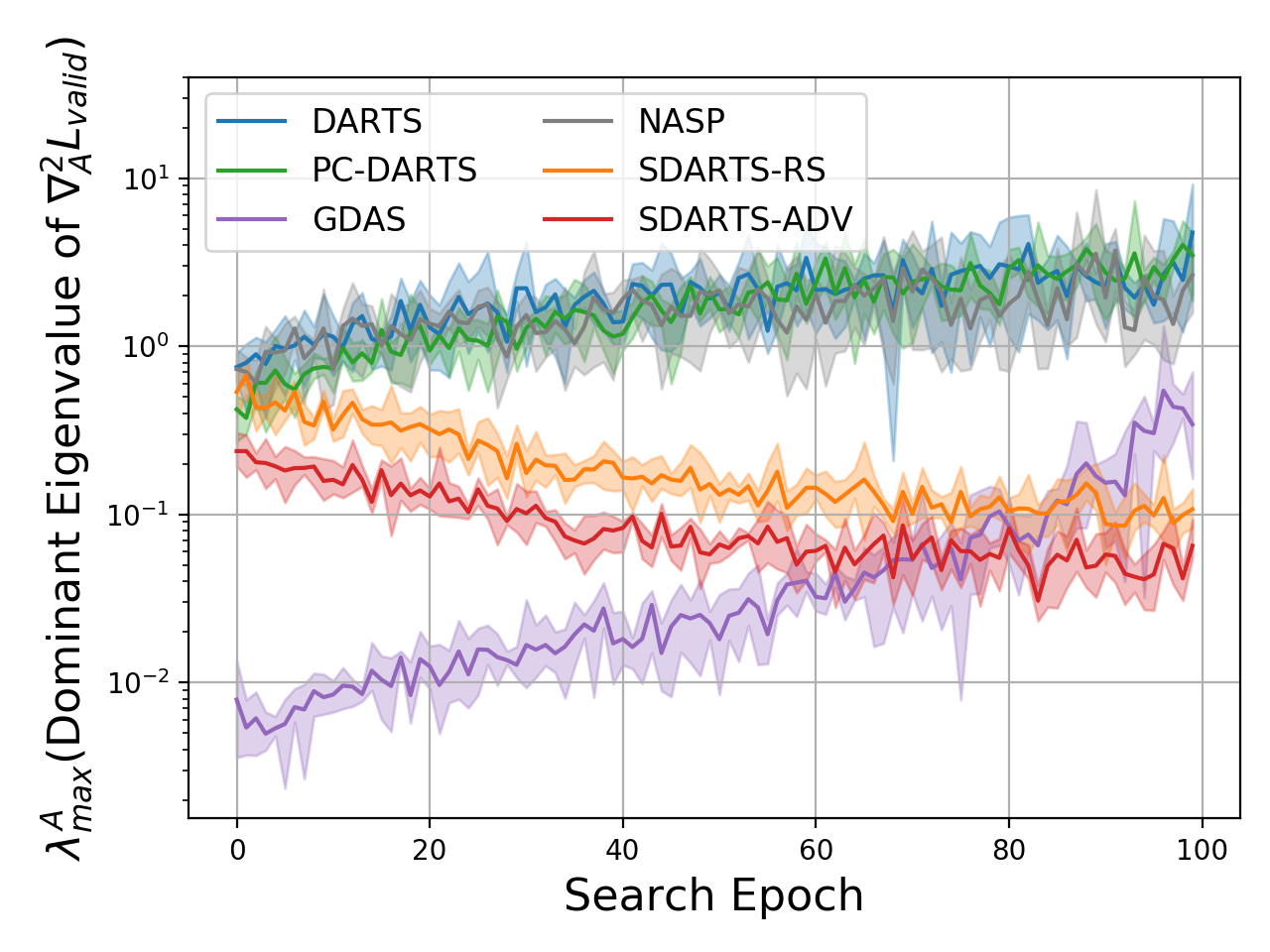}}
\subfigure[Space 3]{\label{fig:space3_eigen}\includegraphics[width=0.33\linewidth]{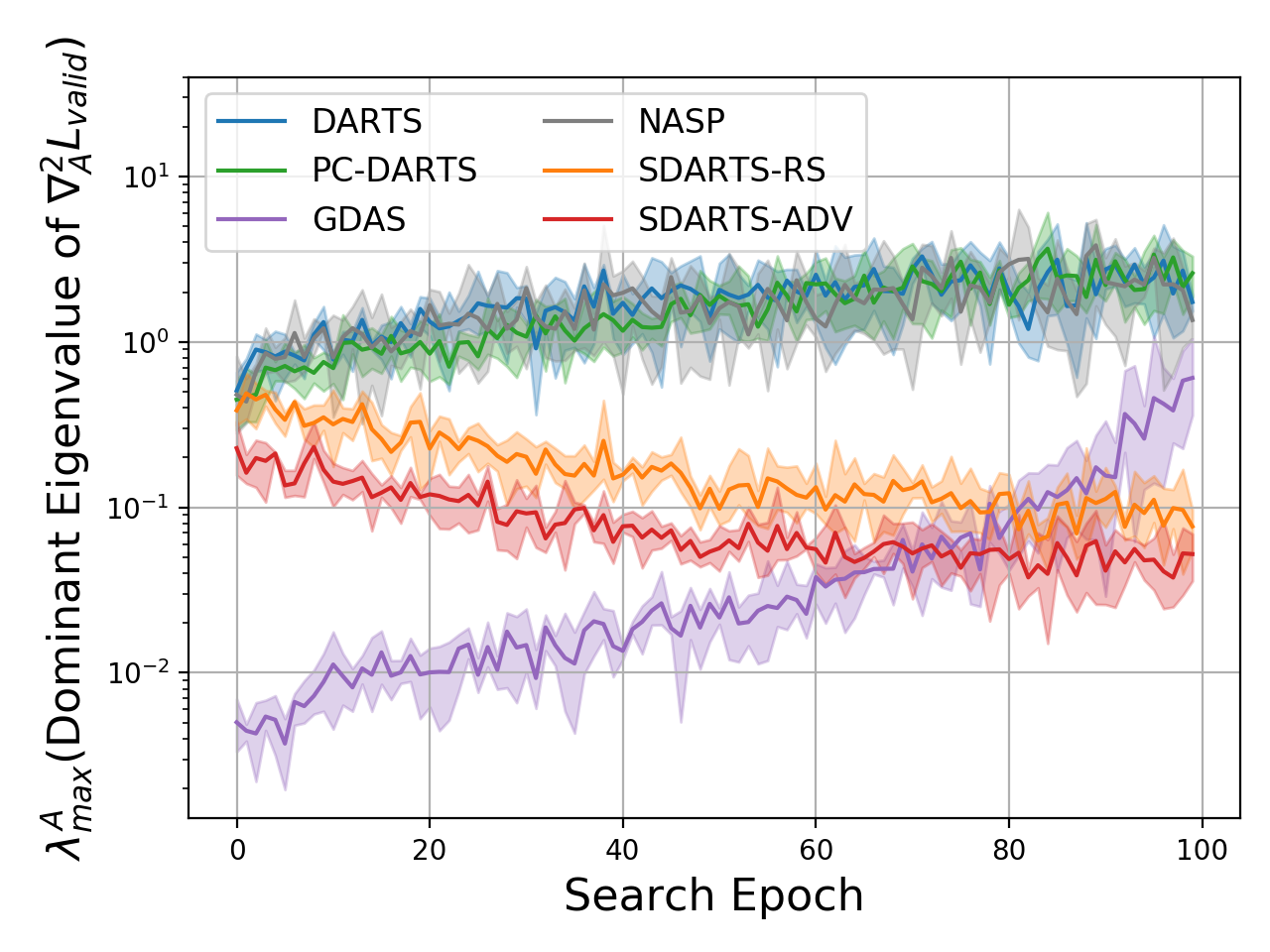}}
}
\caption{Trajectory (mean $\pm$ std) of the Hessian norm on NAS-Bench-1Shot1 (best viewed in color).}
\label{fig:NAS-Bench-1Shot1_eigen}
\end{figure*}

\paragraph{Settings.} 
NAS-Bench-1Shot1 consists of 3 search spaces based on CIFAR-10, which contains 6,240, 29,160 and 363,648 architectures respectively. 
The macro architecture in all spaces is constructed by 3 stacked blocks, with a \textit{max-pooling} operation in between as the DownSampler.
Each block contains 3 stacked cells and the micro architecture of each cell is represented as a DAG.
Apart from the operation on every edge, the search algorithm also needs to determine the topology of edges connecting input, output nodes and the choice blocks.
We refer to their paper \cite{nasbench1shot1} for details of the search spaces.

We make a comparison between our methods with other popular NAS algorithms on all 3 search spaces. Descriptions of the compared baselines can be found in Appendix \ref{app:descript}.
We run every NAS algorithm for 100 epochs (twice of the default DARTS setting) to allow a thorough and comprehensive analysis on search stability and generalizability.
Hyperparameter settings for 5 baselines are set as their defaults.
For both \EXP and \ADV, the perturbation on $A$ is performed after the softmax layer. 
We initialize the norm ball $\epsilon$ as 0.03 and linearly increase it to 0.3 in all our experiments.
The random perturbation $\delta$ in \EXP is sampled uniformly between $-\epsilon$ and $\epsilon$,
and we use the 7-step PGD attack under $\ell_{\infty}$ norm ball to obtain the $\delta$ in \ADV.
Other settings are the same as DARTS.

To search for 100 epochs on a single NVIDIA GTX 1080 Ti GPU, ENAS \cite{enas}, DARTS \cite{darts}, GDAS \cite{gdas}, NASP \cite{nasp}, and PC-DARTS \cite{pcdarts} require 10.5h, 8h, 4.5h, 5h, and 6h respectively.
Extra time of \EXP is just for the random sample, so its search time is approximately the same with DARTS, which is 8h. 
\ADV needs extra steps of forward and backward propagation to perform the adversarial attack, so it spends 16h. 
Notice that this can be largely reduced by setting the PGD attack step as 1 \cite{fgsm}, which brings little performance decrease according to our experiments.

\paragraph{Results.} We plot the anytime test error averaged from 6 independent runs in Figure \ref{fig:NAS-Bench-1Shot1}. 
Also, the trajectory (mean $\pm$ std) of the spectral norm of $\nabla_{A}^{2}L_{valid}$ is shown in Figure \ref{fig:NAS-Bench-1Shot1_eigen}. 
Note that ENAS is not included in Figure \ref{fig:NAS-Bench-1Shot1_eigen} since it does not have the architecture weight $A$.
We provide our detailed analysis below.

\begin{itemize}


\item 
DARTS \cite{darts} generates architectures with deteriorating performance when the search epoch becomes large, which is in accordance with the observations in \cite{understanding}. 
The single-path modifications (GDAS \cite{gdas}, NASP \cite{nasp}) take some effects, e.g. GDAS prevents to find worse architectures and remains stable. 
However, GDAS suffers premature convergence to sub-optimal architectures, and NASP is only effective for the first few search epochs before its performance starts to fluctuate like ENAS. 
PC-DARTS is the best baseline on Space 1 and 3, but it also suffers degenerate performance on Space 2.

\item 
\EXP outperforms all 5 baselines on 3 search spaces.
It better explores the architecture space and meanwhile overcomes the instability issue in DARTS.
\ADV achieves even better performance by forcing $w$ to minimize the worst-case loss around a neighborhood of $A$.
Its anytime test error continues to decrease when the search epoch is larger than 80, which does not occur for any other method.

\item
As explained in Section \ref{sec:RegHessian}, the spectral norm $\lambda_{max}^{A}$ of Hessian $\nabla_{A}^{2}L_{valid}$ has strong correlation with the stability and solution quality. 
Large $\lambda_{max}^{A}$ leads to poor generalizability and stability.
In agreement with the theoretical analysis in Section \ref{sec:RegHessian}, both \EXP and \ADV anneal $\lambda_{max}^{A}$ to a low level throughout the search procedure.
In comparison, $\lambda_{max}^{A}$ in all baselines continue to increase and they even enlarge beyond 10 times after 100 search epochs. 
Though GDAS \cite{gdas} has the lowest $\lambda_{max}^{A}$ at the beginning, it suffers the largest growth rate.
The partial channel connection introduced in PC-DARTS \cite{pcdarts} can not regularize the Hessian norm either, thus PC-DARTS has a similar $\lambda_{max}^{A}$ trajectory to DARTS and NASP, which match their comparably unstable performance.
\end{itemize}

\begin{table*}[!htb]
    \centering
    \captionsetup{justification=centering}
    \caption{Comparison with state-of-the-art image classifiers on CIFAR-10.}
    \resizebox{.7\textwidth}{!}{
    \begin{threeparttable}
    \begin{tabular}{lcccc}
    \hline
    \textbf{Architecture} & \textbf{\tabincell{c}{Mean Test Error\\(\%)}} & \textbf{\tabincell{c}{Params\\(M)}} & \textbf{\tabincell{c}{Search Cost\\(GPU days)}} & \textbf{\tabincell{c}{Search\\Method}} \\ \hline
    DenseNet-BC \cite{densenet}\tnote{$\star$} & 3.46 & 25.6 & - & manual \\ \hline
    
    NASNet-A \cite{nasnet} & 2.65 & 3.3 & 2000 & RL \\
    AmoebaNet-A \cite{amoebanet} & $3.34\pm 0.06$ & 3.2 & 3150 & evolution \\
    AmoebaNet-B \cite{amoebanet} & $2.55\pm0.05$ & 2.8 & 3150 & evolution \\
    PNAS \cite{pnas}\tnote{$\star$} & $3.41\pm 0.09$ & 3.2 & 225 & SMBO \\
    ENAS \cite{enas} & 2.89 & 4.6 & 0.5 & RL \\ 
    NAONet \cite{nao} & 3.53 & 3.1 & 0.4 & NAO \\ \hline
    
    DARTS (1st) \cite{darts} & $3.00\pm 0.14$ & 3.3 & 0.4 & gradient \\
    DARTS (2nd) \cite{darts} & $2.76\pm 0.09$ & 3.3 & 1 & gradient \\
    SNAS (moderate) \cite{snas} & $2.85\pm 0.02$ & 2.8 & 1.5 & gradient \\
    GDAS \cite{gdas} & 2.93 & 3.4 & 0.3 & gradient \\
    BayesNAS \cite{BayesNAS} & $2.81\pm 0.04$ & 3.4 & 0.2 & gradient \\
    ProxylessNAS \cite{proxylessnas}\tnote{$\dagger$} & 2.08 & - & 4.0 & gradient \\
    NASP \cite{nasp} & $2.83\pm 0.09$ & 3.3 & 0.1 & gradient \\
    P-DARTS \cite{pdarts} & 2.50 & 3.4 & 0.3 & gradient \\ 
    PC-DARTS \cite{pcdarts} & $2.57\pm 0.07$ & 3.6 & 0.1 & gradient \\
    R-DARTS(L2) \cite{understanding} & $2.95\pm 0.21$ & - & 1.6 & gradient \\ \hline
    
    \EXP & $2.67\pm 0.03$ & 3.4 & 0.4\tnote{$\ddagger$} & gradient \\
    \ADV & $2.61\pm 0.02$ & 3.3 & 1.3\tnote{$\ddagger$} & gradient \\ 
    PC-DARTS-RS & $2.54\pm 0.04$ & 3.4 & 0.1\tnote{$\ddagger$} & gradient \\
    PC-DARTS-ADV & $2.49\pm 0.04$ & 3.5 & 0.4\tnote{$\ddagger$} & gradient \\
    P-DARTS-RS & $2.50\pm 0.03$ & 3.4 & 0.3\tnote{$\ddagger$} & gradient \\
    P-DARTS-ADV & $2.48\pm 0.02$ & 3.4 & 1.1\tnote{$\ddagger$} & gradient \\
    
    \hline
    \end{tabular}
    
    \begin{tablenotes}
        \item[$\star$] Obtained without cutout augmentation.
        \item[$\dagger$] Obtained on a different space with PyramidNet \cite{pyramidnet} as the backbone.
        \item[$\ddagger$] Recorded on a single GTX 1080Ti GPU.
    \end{tablenotes}
    \end{threeparttable}}
    \label{tab:cifar10}
\end{table*}

\subsection{Experiments on CIFAR-10}
\label{sec:cifar10}
\paragraph{Settings.} 
We employ \EXP and \ADV to search CNN cells on CIFAR-10 following the search space (with 7 possible operations) in DARTS \cite{darts}.
The macro architecture is obtained by stacking convolution cells for 8 times, and every cell contains $N=7$ nodes (2 input nodes, 4 intermediate nodes, and 1 output nodes). Other detailed settings for searching and evaluation can be found in Appendix \ref{app:train_details}, which are the same as DARTS \cite{darts}.
To further demonstrate the effectiveness of the proposed regularization, we also test our methods on popular DARTS variants PC-DARTS \cite{pcdarts} (shown as PC-DARTS-RS and PC-DARTS-ADV) and P-DARTS \cite{pdarts} (shown as P-DARTS-RS and P-DARTS-ADV).

\paragraph{Results.} 
Table \ref{tab:cifar10} summarizes the comparison of our methods with state-of-the-art algorithms, and the searched normal cells are visualized in Figure \ref{fig:normal}. 
Compared with the original DARTS, the random smoothing regularization (SDARTS-RS) decreases the test error from 3.00\% to 2.67\%, and the adversarial regularization (SDARTS-ADV) further decreases it to 2.61\%.
When applying to PC-DARTS and P-DARTS, both regularization techniques achieve consistent performance gain and obtain highly competitive results.
We also reduces the variance of the search result.


\subsection{Experiments on ImageNet}
\label{sec:imagenet}
\paragraph{Settings.} 
We test the transferability of our discovered cells on ImageNet. Here the network is constructed by 14 cells and 48 initial channels. We train the network for 250 epochs by an SGD optimizer with an annealing learning rate initialized as 0.5, a momentum of 0.9, and a weight decay of $3\times 10^{-5}$. 
Similar to previous works \cite{pcdarts,pdarts}, we also employ label smoothing and auxiliary loss tower to enhance the training. 

\paragraph{Results.} 
As shown in Table \ref{tab:imagenet}, both \EXP and \ADV outperform DARTS by a large margin.
Moreover, both regularization methods achieve improved accuracy when applying to PC-DARTS and P-DARTS, which demonstrates their generalizability and effectiveness on large-scale tasks. Our best run achieves a top1/5 test error of 24.2\%/7.2\%, ranking top amongst popular NAS methods.

\begin{table}[!htb]
    \centering
    \caption{Comparison with state-of-the-art image classifiers on ImageNet in the mobile setting.}
    \resizebox{.45\textwidth}{!}{
    \begin{tabular}{lcc}
    \hline
    
    \multirow{2}*{\textbf{Architecture}} & \multicolumn{2}{c}{\textbf{Test Error(\%)}} \\ \cline{2-3}
    & top-1 & top-5 \\ \hline
    
    Inception-v1 \cite{inception-v1} & 30.1 & 10.1 \\
    MobileNet \cite{mobilenets} & 29.4 & 10.5 \\
    ShuffleNet (v1) \cite{shufflenet-v1} & 26.4 & 10.2 \\
    ShuffleNet (v2) \cite{shufflenet-v2} & 25.1 & 10.1 \\ \hline
    
    NASNet-A \cite{nasnet} & 26.0 & 8.4 \\
    AmoebaNet-C \cite{amoebanet} & 24.3 & 7.6 \\
    PNAS \cite{pnas} & 25.8 & 8.1 \\
    MnasNet-92 \cite{mnasnet} & 25.2 & 8.0 \\ \hline
    
    DARTS \cite{darts} & 26.7 & 8.7 \\
    SNAS (mild) \cite{snas} & 27.3 & 9.2 \\
    GDAS \cite{gdas} & 26.0 & 8.5 \\
    ProxylessNAS (GPU) \cite{proxylessnas} & 24.9 & 7.5 \\
    NASP \cite{nasp} & 27.2 & 9.1 \\
    P-DARTS \cite{pdarts} & 24.4 & 7.4 \\ 
    PC-DARTS \cite{pcdarts} & 25.1 & 7.8 \\ \hline
    
    SDARTS-RS & 25.6 & 8.2 \\ 
    SDARTS-ADV & 25.2 & 7.8 \\ 
    PC-DARTS-RS & 24.7 & 7.5 \\ 
    PC-DARTS-ADV & 24.3 & 7.4\\
    P-DARTS-RS & 24.4 & 7.4\\ 
    P-DARTS-ADV & 24.2 & 7.2 \\ \hline
    
    \end{tabular}}
    \label{tab:imagenet}
\end{table}

\subsection{Experiments on PTB}
\label{sec:rnn_standard}
\paragraph{Settings.}
Besides searching for CNN cells, our methods are applicable to various scenarios such as identifying RNN cells. 
Following DARTS \cite{darts}, the RNN search space based on PTB contains 5 candidate functions, \textit{tanh}, \textit{relu}, \textit{sigmoid}, \textit{identity} and \textit{zero}. 
The macro architecture of the RNN network is comprised of only a single cell consisting of $N=12$ nodes. 
The first intermediate node is manually fixed and the rest nodes are determined by the search algorithm.
When searching, we train the RNN network for 50 epochs with sequence length as 35. 
During evaluation, the final architecture is trained by an SGD optimizer, where the batch size is set as 64 and the learning rate is fixed as 20. These settings are the same as DARTS.

\paragraph{Results.}
The results are shown in Table \ref{tab:rnn_standard}. 
\EXP achieves a validation perplexity of 58.7 and a test perplexity of 56.4. Meanwhile, \ADV achieves a validation perplexity of 58.3 and a test perplexity of 56.1. 
We outperform other NAS methods with similar model size, which demonstrates the effectiveness of our methods for the RNN space.
LSTM + SE \cite{lstm} obtains better results than us, but it benefits from a handcrafted ensemble structure.


\begin{table}[!htb]
    \centering
    \caption{Comparison with state-of-the-art language models on PTB (lower perplexity is better).}
    \resizebox{.48\textwidth}{!}{
    \begin{threeparttable}
    \begin{tabular}{lcccc}
    \hline
    \multirow{2}*{\textbf{Architecture}} & \multicolumn{2}{c}{\textbf{Perplexity(\%)}} & \multirow{2}*{\textbf{\tabincell{c}{Params\\(M)}}} \\ \cline{2-3}
    & valid & test & \\ \hline
    LSTM + SE \cite{lstm}\tnote{$\star$} & 58.1 & 56.0 & 22 \\ \hline
    NAS \cite{nas} & - & 64.0 & 25 \\
    ENAS \cite{enas} & 60.8 & 58.6 & 24 \\ \hline
    DARTS (1st) \cite{darts} & 60.2 & 57.6 & 23 \\
    DARTS (2nd) \cite{darts}\tnote{$\dagger$} & 58.1 & 55.7 & 23 \\
    GDAS \cite{gdas} & 59.8 & 57.5 & 23 \\
    NASP \cite{nasp} & 59.9 & 57.3 & 23 \\ \hline
    \EXP & 58.7 & 56.4 & 23 \\
    \ADV & 58.3 & 56.1 & 23 \\ \hline
    \end{tabular}
    \begin{tablenotes}
        \item[$\star$] LSTM + SE represents LSTM with 15 softmax experts.
        \item[$\dagger$] We achieve 58.5 for validation and 56.2 for test when training the architecture found by DARTS (2nd) ourselves.
    \end{tablenotes}
    \end{threeparttable}
    }
    \label{tab:rnn_standard}
\end{table}

\begin{table*}[!htb]
    \centering
    \captionsetup{justification=centering}
    \caption{Comparison with popular regularization techniques (test error (\%)). \\ The best method is boldface and underlined while the second best is boldface.}
    \resizebox{.84\textwidth}{!}{
    \begin{tabular}{c|c|c|c|c|c|c|c|c}
    \hline
    \textbf{Dataset} & \textbf{Space} & \textbf{DARTS} & \textbf{PC-DARTS} & \textbf{DARTS-ES} & \textbf{R-DARTS(DP)} & \textbf{R-DARTS(L2)} & \textbf{\EXP} & \textbf{\ADV} \\ \hline
    \multirow{4}*{C10} & S1 & 3.84 & 3.11 & 3.01 & 3.11 & \textbf{2.78} & \textbf{2.78} & \underline{\textbf{2.73}} \\
    & S2 & 4.85 & 3.02 & 3.26 & 3.48 & 3.31 & \textbf{2.75} & \underline{\textbf{2.65}} \\
    & S3 & 3.34 & \textbf{2.51} & 2.74 & 2.93 & \textbf{2.51} & 2.53 & \underline{\textbf{2.49}} \\
    & S4 & 7.20 & 3.02 & 3.71 & 3.58 & 3.56 & \textbf{2.93} & \underline{\textbf{2.87}} \\ \hline
    
    \multirow{4}*{C100} & S1 & 29.46 & 24.69 & 28.37 & 25.93 & 24.25 & \textbf{23.51} & \underline{\textbf{22.33}} \\
    & S2 & 26.05 & 22.48 & 23.25 & 22.30 & 22.44 & \textbf{22.28} & \underline{\textbf{20.56}} \\
    & S3 & 28.90 & 21.69 & 23.73 & 22.36 & 23.99 & \textbf{21.09} & \underline{\textbf{21.08}} \\
    & S4 & 22.85 & 21.50 & \textbf{21.26} & 22.18 & 21.94 & 21.46 & \underline{\textbf{21.25}} \\ \hline
    
    \multirow{4}*{SVHN} & S1 & 4.58 & 2.47 & 2.72 & 2.55 & 4.79 & \textbf{2.35} & \underline{\textbf{2.29}} \\
    & S2 & 3.53 & 2.42 & 2.60 & 2.52 & 2.51 & \textbf{2.39} & \underline{\textbf{2.35}} \\
    & S3 & 3.41 & 2.41 & 2.50 & 2.49 & 2.48 & \underline{\textbf{2.36}} & \textbf{2.40} \\
    & S4 & 3.05 & \textbf{2.43} & 2.51 & 2.61 & 2.50 & 2.46 & \underline{\textbf{2.42}} \\ \hline
    \end{tabular}}
    \label{tab:compare_reg}
\end{table*}

\subsection{Comparison with Other Regularization}
\label{sec:compare_reg}
Our methods can be viewed as a way to regularize DARTS (implicitly regularize the Hessian norm of validation loss).
In this section, we compare \EXP and \ADV with other popular regularization techniques.
The compared baselines are 1) partial channel connection (PC-DARTS \cite{pcdarts}); 2) ScheduledDropPath \cite{nasnet} (R-DARTS(DP)); 3) L2 regularization on $w$ (R-DARTS(L2)); 3) early stopping (DARTS-ES \cite{understanding}). 
Descriptions of the compared regularization baselines are shown in Appendix \ref{app:descript}.

\paragraph{Settings.} 
We perform a thorough comparison on 4 simplified search spaces proposed in \cite{understanding} across 3 datasets (CIFAR-10, CIFAR-100, and SVHN).
All simplified search spaces only contain a portion of candidate operations (details are shown in Appendix \ref{app:4space}).
Following \cite{understanding}, we use 20 cells with 36 initial channels for CIFAR-10, and 8 cells with 16 initial channels for CIFAR-100 and SVHN.
The rest settings are the same with Section \ref{sec:cifar10}.
Results in Table \ref{tab:compare_reg} are obtained by running every method 4 independent times and pick the final architecture based on the validation accuracy (retrain from scratch for a few epochs).

\paragraph{Results.}
Our methods achieve consistent performance gain compared with baselines.
\ADV is the best method for 11 out of 12 benchmarks and we take over both first and second places for 9 benchmarks.
In particular, \ADV outperforms DARTS, R-DARTS(L2), DARTS-ES, R-DARTS(DP), and PC-DARTS by 31.1\%, 11.5\%, 11.4\%, 10.9\%, and 5.3\% on average.

\subsection{Examine the Searched Architectures}
\label{sec:examine_arch}
As pointed out in \cite{understanding, WideShallow}, DARTS tends to fall into distorted architectures that converge faster, which is another manifestation of its instability.
So here we examine the generated architectures and see whether our methods can overcome such bias.

\subsubsection{Proportion of Parameter-Free Operations}
Many \cite{understanding,darts+} have found out that parameter-free operations such as \textit{skip connection} dominate the generated architecture.
Though makes architectures converge faster, excessive parameter-free operations can largely reduce the model's representation capability and bring out low test accuracy.
As illustrated in Table \ref{tab:proportion_op}, we also find similar phenomenon when searching by DARTS on 4 simplified search spaces in Section \ref{sec:compare_reg}. 
The proportion of parameter-free operations even becomes 100\% on S1 and S3, and DARTS can not distinguish the harmful \textit{noise} operation on S4.
PC-DARTS achieves some improvements but is not enough since \textit{noise} still appears.
DARTS-ES reveals its effectiveness on S2 and S4 but fails on S3 since all operations found are \textit{skip connection}.
We do not show R-DARTS(DP) and R-DARTS(L2) here because their discovered cells are not released.
In comparison, both \EXP and \ADV succeed in controlling the portion of parameter-free operations on all search spaces.


\subsubsection{Connection Pattern}
\citet{WideShallow} demonstrates that DARTS favors wide and shallow cells since they often have smoother loss landscape and faster convergence speed.
However, these cells may not generalize better than their narrower and deeper variants \cite{WideShallow}.
Follow their definitions (suppose every intermediate node has width $c$, detailed definitions are shown in Appendix \ref{app:def_width_height}), the best cell generated by our methods on CNN standard space (Section \ref{sec:cifar10}) has width 3$c$ and depth 4. 
In contrast, ENAS has width 5$c$ and depth 2, DARTS has width 3.5$c$ and depth 3, PC-DARTS has width 4$c$ and depth 2.
Consequently, we succeed in mitigating the bias of connection pattern.

\begin{table}[!h]
    \centering
    \caption{Proportion of parameter-free operations in normal cells found on CIFAR-10.}
    \resizebox{0.48\textwidth}{!}{
    \begin{tabular}{c|c|c|c|c|c}
    \hline
    \textbf{Space} & \textbf{DARTS} & \textbf{PC-DARTS} & \textbf{DARTS-ES} & \textbf{\EXP} & \textbf{\ADV} \\ \hline
    S1 & 1.0 & 0.5 & 0.375 & 0.125 & 0.125 \\ 
    S2 & 0.875 & 0.75 & 0.25 & 0.375 & 0.125 \\
    S3 & 1.0 & 0.125 & 1.0 & 0.125 & 0.125 \\
    S4 & 0.625 & 0.125 & 0.0 & 0.0 & 0.0 \\ \hline
    \end{tabular}}
    \label{tab:proportion_op}
\end{table}

\section{Conclusion}
We introduce SmoothDARTS (SDARTS), a perturbation-based regularization to improve the stability and generalizability of differentiable architecture search. 
Specifically, the regularization is carried out with random smoothing or adversarial attack.
SDARTS possesses a much smoother landscape and has the theoretical guarantee to regularize the Hessian norm of the validation loss.
Extensive experiments illustrate the effectiveness of SDARTS and we outperform various regularization techniques.

\section*{Acknowledgement}
We thank Xiaocheng Tang, Xuanqing Liu, and Minhao Cheng for the constructive discussions. This work is supported by NSF IIS-1719097, Intel, and Facebook. 

\bibliography{bib}
\bibliographystyle{icml2020}

\clearpage

\section{Appendix}


\subsection{Descriptions of compared baselines}
\label{app:descript}

\begin{itemize}
\item \textbf{ENAS} \cite{enas} first trains the shared parameter of a one-shot network. For the search phase, it samples sub-networks and use the validation error as the reward signal to update an RNN controller following REINFORCE \cite{reinforce} rule. Finally, they sample several architectures guided by the trained controller and derive the one with the highest validation accuracy.

\item \textbf{DARTS} \cite{darts} builds a mixture architecture similar to ENAS. The difference is that it relaxes the discrete architecture space to a continuous and differentiable representation by assigning a weight $\alpha$ to every operation. The network weight $w$ and $\alpha$ are then updated via gradient descent alternately based on the training set and the validation set respectively. For evaluation, DARTS prunes out all operations except the one with the largest $\alpha$ on every edge, which leaves the final architecture.

\item \textbf{GDAS} \cite{gdas} uses the Gumbel-Softmax trick to activate only one operation for every edge during search, similar technique is also applied in SNAS \cite{snas}. This trick reduces the memory cost during search meanwhile keeps the property of differentiability.

\item \textbf{NASP} \cite{nasp} is another modification of DARTS via the proximal algorithm. 
A discrete version of architecture weight $\bar{A}$ is computed every search epoch by applying a proximal operation to the continuous $A$. Then the gradient of $\bar{A}$ is utilized to update its corresponding $A$ after backpropagation.

\item \textbf{PC-DARTS} \cite{pcdarts} evaluates only a random proportion of the channels. This partial channel connection not only accelerates search but also serves as a regularization that controls the bias towards parameter-free operations, as explained by the author.

\item \textbf{R-DARTS(DP)} \cite{understanding} runs DARTS with different intensity of ScheduledDropPath regularization \cite{nasnet} and picks the final architecture according to the performance on the validation set. In ScheduledDropPath, each path in the cell is dropped out with a probability that increases linearly over the training procedure.

\item \textbf{R-DARTS(L2)} \cite{understanding} runs DARTS with different amounts of L2 regularization and selects the final architecture in the same way with R-DARTS(DP). Specifically, the L2 regularization is applied on the inner loop (i.e. network weight $w$) of the bi-level optimization problem.

\item \textbf{DARTS-ES} \cite{understanding} early stops the search procedure of DARTS if the increase of $\lambda_{max}^{A}$ (the dominate eigenvalue of Hessian $\nabla_{A}^{2}L_{valid}$) exceeds a threshold. This prevents $\lambda_{max}^{A}$, which is highly correlated with the stability and generalizability of DARTS, from exploding.

\end{itemize}

\subsection{Training details on CNN standard space}
\label{app:train_details}
For the search phase, we train the mixture architecture for 50 epochs, with the 50K CIFAR-10 dataset be equally split into training and validation set.
Following \cite{darts}, the network weight $w$ is optimized on the training set by an SGD optimizer with momentum as 0.9 and weight decay as $3\times 10^{-4}$, where the learning rate is annealed from 0.025 to 1e-3 following a cosine schedule. 
Meanwhile, we use an Adam optimizer with learning rate 3e-4 and weight decay 1e-3 to learn the architecture weight $A$ on the validation set. 
For the evaluation phase, the macro structure consists of 20 cells and the initial number of channels is set as 36. 
We train the final architecture by 600 epochs using the SGD optimizer with a learning rate cosine scheduled from 0.025 to 0, a momentum of 0.9 and a weight decay of 3e-4.
The drop probability of ScheduledDropPath increases linearly from 0 to 0.2, and the auxiliary tower \cite{nas} is employed with a weight of 0.4.
We also utilize CutOut \cite{cutout} as the data augmentation technique and report the result (mean $\pm$ std) of 4 independent runs with different random seeds.

\subsection{Micro architecture of 4 simplified search spaces}
\label{app:4space}
\begin{figure*}[!htb]
\centering
\subfigure[Normal cell]{\label{fig:normal_s1}\includegraphics[width=1.0\linewidth]{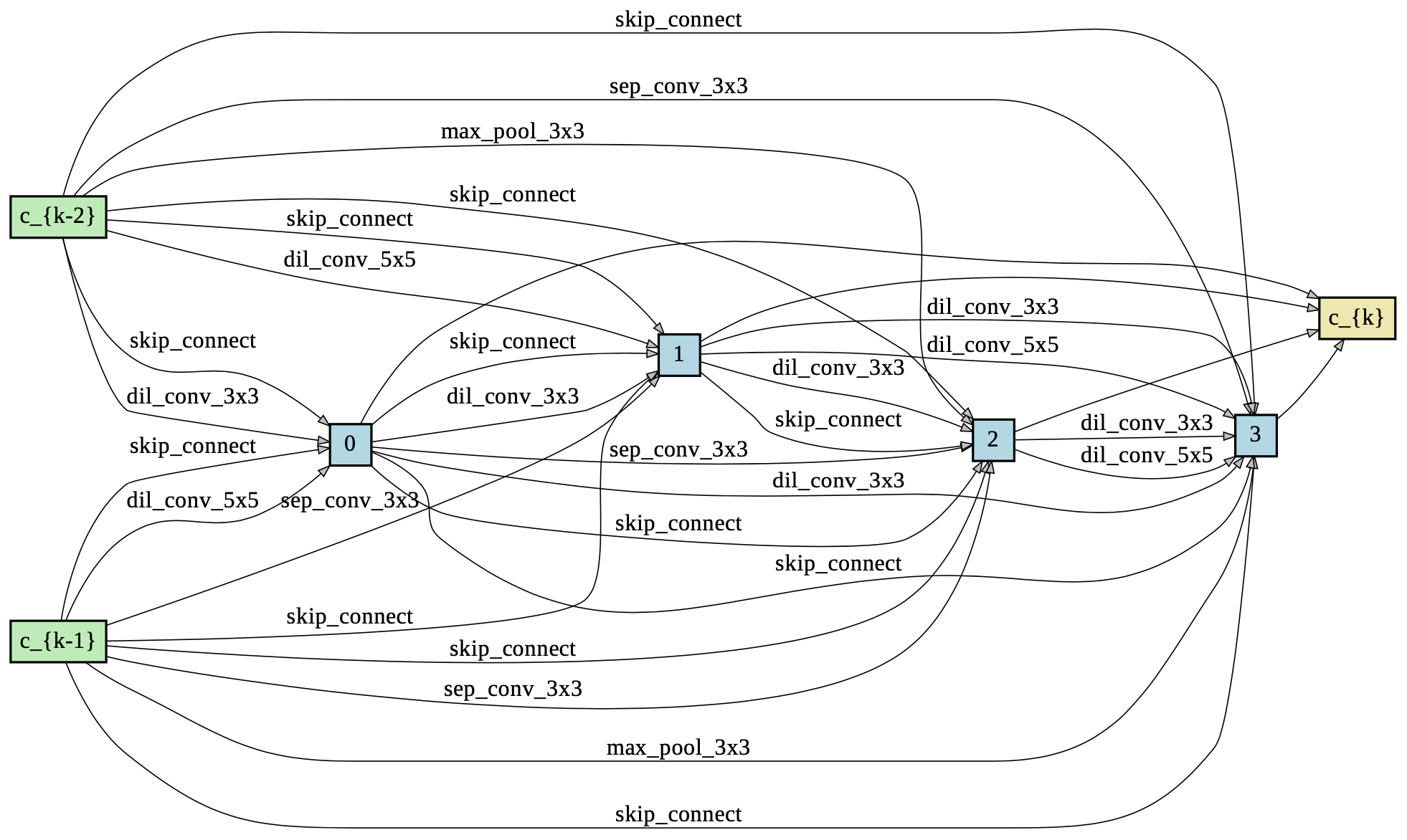}}
\subfigure[Reduction cell]{\label{fig:reduction_s1}\includegraphics[width=1.0\linewidth]{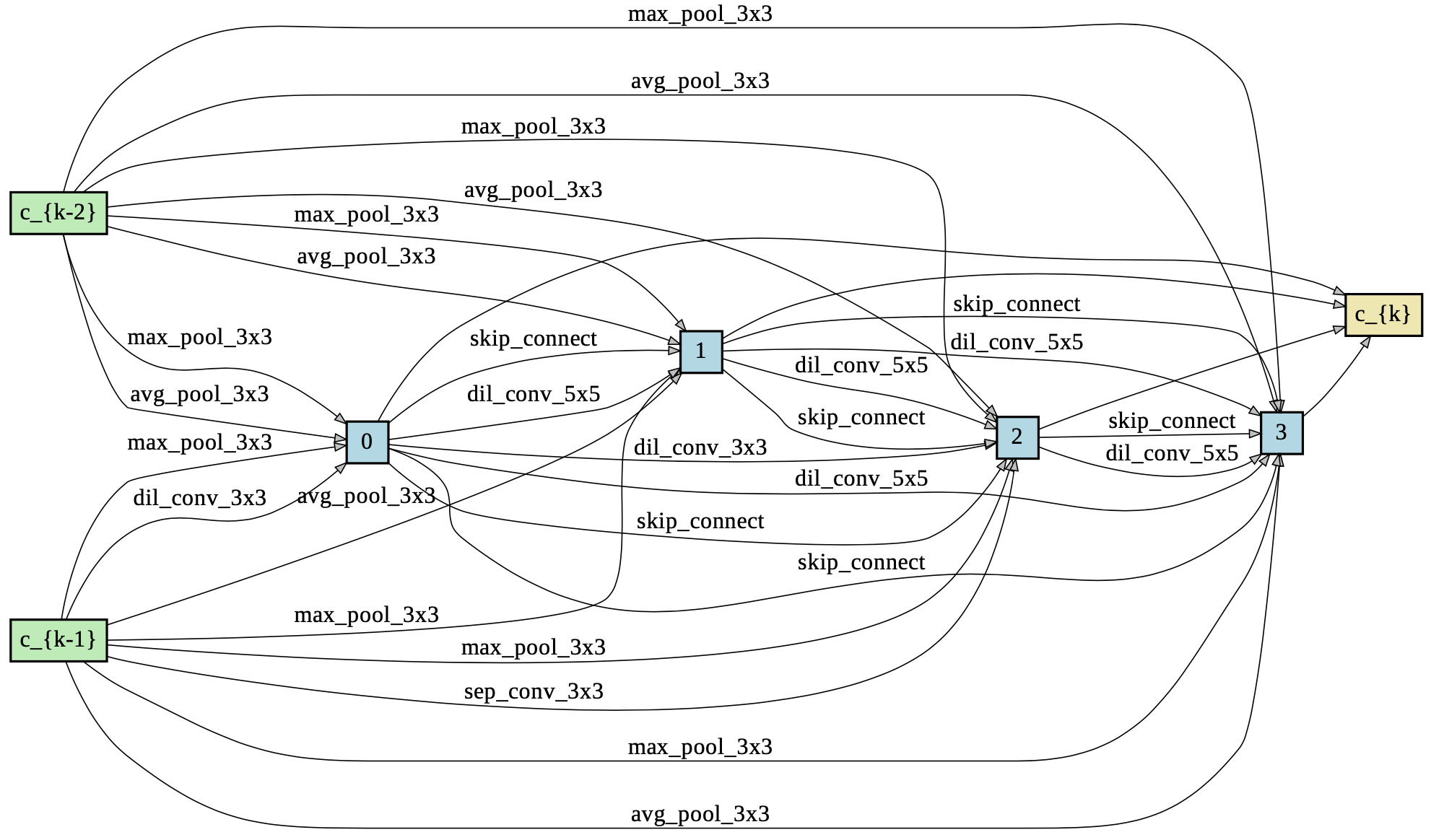}}
\caption{Micro cell architecture of S1.}
\label{fig:s1}
\end{figure*}

The first space S1 contains 2 popular operators per edge as shown in Figure \ref{fig:s1}, S2 restricts the set of candidate operations on every edge as \{$3\times 3$ \textit{separable convolution}, \textit{skip connection}\}, the operation set in S3 is \{$3\times 3$ \textit{separable convolution}, \textit{skip connection}, \textit{zero}\}, and S4 simplifies the set as \{$3\times 3$ \textit{separable convolution}, \textit{noise}\}.

\subsection{Definitions of cell width and height}
\label{app:def_width_height}
Specifically, the depth of a cell is the number of connections on the longest path from input nodes to the output node.
While the width of a cell is computed by adding the width of all intermediate nodes that are directly connected to the input nodes, where the width of a node is defined as the channel number for convolutions and the feature dimension for linear operations (In \cite{WideShallow}, they assume the width of every intermediate node is $c$ for simplicity). 
In particular, if an intermediate node is partially connected to input nodes (i.e. has connections to other intermediate nodes), its width is deducted by the percentage of intermediate nodes it is connected to when computing the cell width.

\end{document}